\documentclass{article} 
\usepackage{iclr2025_conference,times}
\usepackage{graphicx}
\usepackage{enumitem}
\usepackage[utf8]{inputenc}
\usepackage{tcolorbox}
\usepackage{pifont} 

\tcbuselibrary{listingsutf8}

\usepackage{amsmath,amsfonts,bm}

\def\eqref#1{equation~\ref{#1}}

\def\1{\bm{1}}

\DeclareMathAlphabet{\mathsfit}{\encodingdefault}{\sfdefault}{m}{sl}
\SetMathAlphabet{\mathsfit}{bold}{\encodingdefault}{\sfdefault}{bx}{n}

\usepackage{hyperref}
\usepackage{url}

\iclrfinalcopy

\title{MedicalBench: Evaluating Large Language Models \\
Towards Improved Medical Concept Extraction}

\author{
\and
Zhichao Yang\\
Optum AI \and 
Gregory D. Lyng$^{\ast}$\\
Optum AI \and 
Sanjit Singh Batra$^{\ast}$\\
Optum AI \and 
Robert E. Tillman\thanks{Correspondence to: \texttt{{{gregory.lyng, sanjit.batra, rob.tillman}} @optum.com}}\\
Optum AI
}

\begin{document}

\maketitle

\begin{abstract}

Medical concept extraction from electronic health records underpins many downstream applications, yet remains challenging because medically meaningful concepts, such as diagnoses, are frequently implied rather than explicitly stated in medical narratives. Existing benchmarks with human-annotated evidence spans underscore the importance of grounding extracted concepts in medical text. However, they predominantly focus on explicitly stated concepts and provide limited coverage of cases in which medically relevant concepts must be inferred.
We present MedicalBench, a new benchmark for medical concept extraction with evidence grounding that evaluates implicit medical reasoning. MedicalBench formulates medical concept extraction as a verification task over medical note concept pairs, coupled with sentence level evidence identification. Built from MIMIC-IV discharge summaries and human verified ICD-10 codes, the dataset is curated through a multi stage large language model (LLM) triage pipeline followed by medical annotation and expert review. It deliberately includes implicit positives, semantically confusable negatives, and cases where LLM judgments disagree with medical expert assessments. Annotators provide sentence level evidence spans and concise medical rationales. The final dataset contains 823 high quality examples.
We define two complementary evaluation tasks: (1) medical concept extraction and (2) sentence level evidence retrieval, enabling assessment of both correctness and interpretability. Benchmarking state-of-the-art LLMs and a supervised baseline reveals that performance remains modest, highlighting the difficulty of extracting implicitly expressed concepts. We further show that explicitly incorporating reasoning cues and prompting to extract implicit evidence substantially improves medical concept extractions, while performance is largely invariant to note length, indicating that MedicalBench isolates reasoning difficulty rather than superficial confounders.
MedicalBench provides the first systematic benchmark for implicit, evidence-grounded medical concept extraction, offering a foundation for developing medical language models that can both identify medically relevant concepts and justify their predictions in a transparent and medically faithful manner.\footnote{MedicalBench dev data is available at \href{https://physionet.org/content/mimic-iv-ext-medicalbench/1.0.0/}{PhysioNet}. MedicalBench test benchmark is available at \href{https://www.codabench.org/competitions/18134/}{Codabench}.}
\end{abstract}

\section{Introduction}

\begin{figure}[h]
\begin{center}
\includegraphics[width=0.90\textwidth]{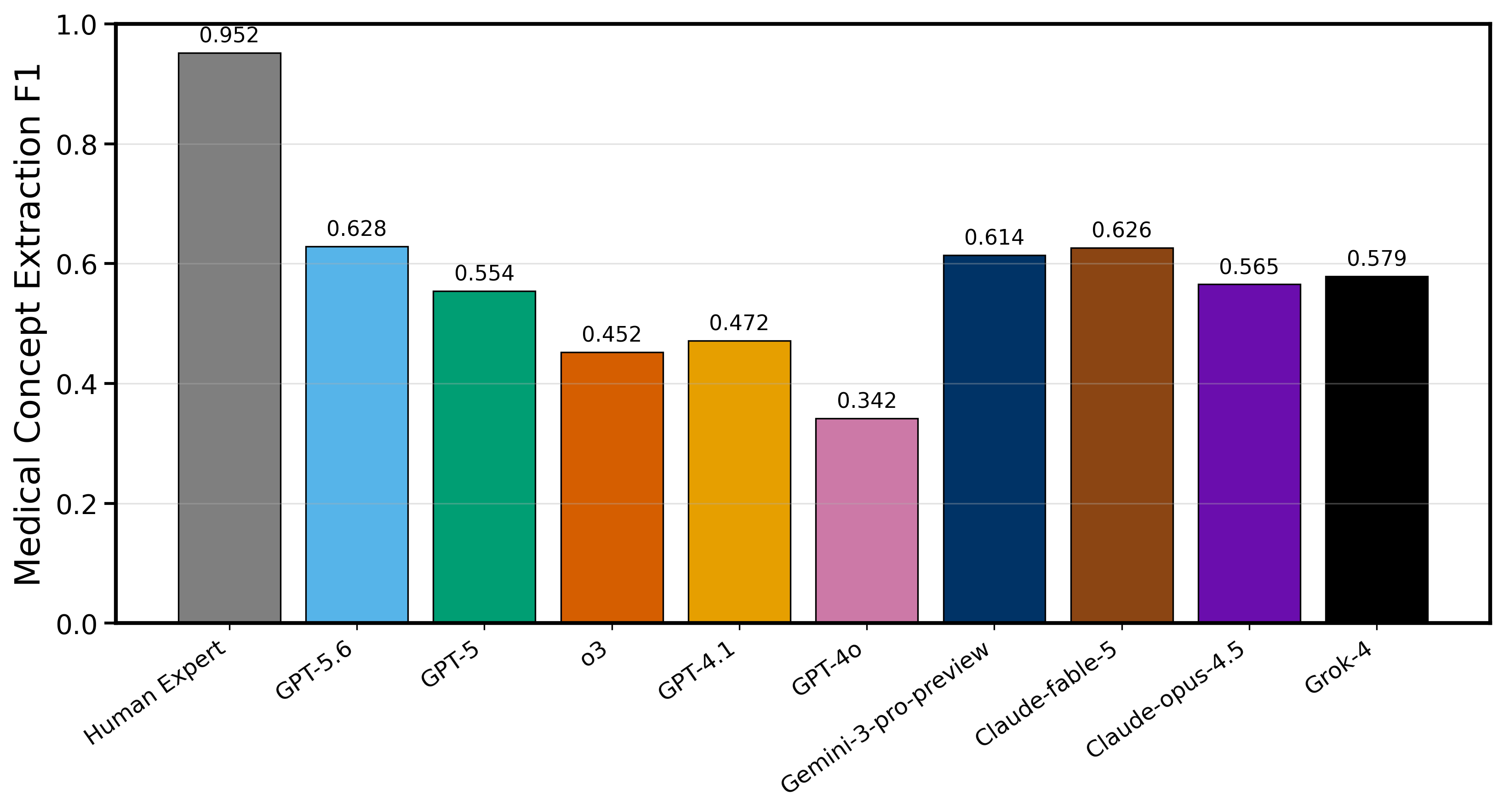}
\end{center}
\caption{Medical concept extraction F1 across medical experts and LLMs on our MedicalBench.
}
\label{fig:main}
\end{figure}

Medical concept extraction aims to identify and structure key medical facts from unstructured clinical narratives such as discharge summaries, progress notes, and oncology reports \citep{wiest2025software,wu2024clinical,sushil2024coral,Landolsi2022InformationEF}. Accurate extraction of diagnostic events is fundamental for building reliable clinical knowledge systems and it powers numerous downstream applications \citep{Huang2024ACA,goldstein2016opportunities}. For example, converting free-text notes into structured diagnoses and findings can aid cohort identification for research \citep{Majid2025Practice,hein2025iterative,Sarmiento2016ImprovingPC}, improve clinical outcome prediction \citep{sun2025evaluating,seinen2022use,shukla2020integrating}, and streamline automated documentation processes\citep{shahid2025using,das2025weakly,Lau2021EventBasedCF}. Robust medical concept extraction underpins many secondary uses of electronic health records (EHRs) by making implicit information explicit and machine-readable \citep{gao2024optimising, tsai2020effects, Xiao2018EHR}.

A core challenge in medical concept extraction is that medical concept related information in narrative text is often implicit, fragmented, and context-dependent \citep{perera-etal-2015-implicit,meystre2008extracting}. For instance, a note might report a ``low hemoglobin level'' as a finding (suggesting concept anemia) or list a body mass index ``BMI 37'' (indicating concept obesity) without explicitly using the words anemia or obesity. Empirical studies have shown that such implicit documentation is widespread: although more than half of patients may meet objective criteria for obesity, only a small fraction have the condition explicitly recorded in structured problem lists \citep{Mattar2017ThePO}.

One prominent line of work that exposes this limitation frames medical concept extraction as a verification and grounding problem: rather than merely identifying which concepts may apply to a patient, automated extraction models must determine whether each candidate concept is supported by the clinical narrative and identify the corresponding evidence. Recent datasets with expert-annotated evidence spans demonstrate the importance of grounding extracted concepts in text, but they remain limited on explicitly stated concepts and offer limited coverage of cases where concepts must be infered from indirect evidences \citep{deyoung2022entity,cheng-etal-2023-mdace,beckh-etal-2025-anatomy}. 
More broadly, most existing benchmarks emphasize concept presence or absence at the document level, while providing little insight into why a concept is warranted or where the supporting evidence resides \citep{mullenbach-etal-2018-explainable}.

In this work, we introduce MedicalBench, a benchmark designed to evaluate medical concept extraction with explicit evidence grounding in realistic medical notes. 
MedicalBench formulates the task as follows: given a medical note and a candidate medical concept, the model must determine whether the concept is supported by the text and, if so, locate the relevant evidence spans. Using discharge summaries from MIMIC-IV and medical concept labels, we construct challenging positive and negative examples through a multi-stage large language model (LLM) triage pipeline followed by rigorous expert annotation. The dataset emphasizes implicit cases and semantically confusable negatives, with expert annotators marking supporting evidence and providing explanatory rationales. To ensure annotation quality, we incorporate control labels derived from LLM-based heuristics to assess annotator reliability.
The resulting dataset comprises 823 high-quality, expert-annotated examples and, to our knowledge, represents the first benchmark to systematically evaluate implicit medical concept extraction with verified evidence grounding.
Across a range of large language models, we find that performance remains limited, with the highest extraction F1 score below 0.7 (Figure~\ref{fig:main}), underscoring the difficulty of this task.
MedicalBench provides a foundation for the development and evaluation of medical language models that can both identify medically meaningful concepts and justify their predictions in a transparent and medically faithful manner.

\section{Data collection}

\subsection{Data Source and Candidate Concepts}
We constructed the dataset using the MIMIC-IV database, which contains de-identified electronic health records of patients admitted to the Beth Israel Deaconess Medical Center \citep{johnson2023mimic}, through both MIMIC-IV repository \citep{johnson2020mimic} and MIMIC-IV-Note repository \citep{johnson2023mimicnote} from physionet \citep{physiobank2000physionet}.
Following \citet{mullenbach-etal-2018-explainable}, we extracted the discharge summary as the representative medical note for each hospital admission (`hadm\_id`). 
Each admission was then mapped to all associated ICD-10 diagnoses in the `diagnoses\_icd` table and ICD-10 procedures in the `procedures\_icd` table. 
Because these billing codes are assigned and reviewed by professional medical coders, we treated them as the clinical concepts of interest. 
Alongside each code, we preserved metadata including the ICD chapter, parent hierarchy, and textual description for prompting and analysis.

\subsection{LLM-based Triage for Positive Candidates}

To identify challenging cases where LLMs may disagree with human experts, we performed a two-stage triage process. In the first stage, a non-reasoning LLM (GPT-4o-mini) was prompted with the discharge summary and the candidate medical concepts, and asked to classify each note–concept pair into one of three categories: Explicit (the concept is explicitly supported by text spans in the note), Implicit (the concept is supported but requires clinical inference, e.g., ``low hemoglobin level'' implying anemia), or Unrelated (the concept is not supported). Each pair is prompted 3 times, and the final answer is achieved by majority vote. Explanations and evidence spans (character offsets) were stored internally. The specific prompt is shown in Section \ref{sec:prompt}.

In the second stage, previous disagreements were re-evaluated using a reasoning LLM (o3). We retained only those cases where both LLMs judged the concept as ``Unrelated'', even though the concept (ICD-10 code) was assigned in MIMIC-IV for that admission. This subset (putative positive) represents candidate false negatives for LLMs and forms the basis for challenging positive examples.

\subsection{Negative Sampling Strategy}

To construct difficult negative examples, we designed two complementary sampling strategies.

\textbf{Prevalence-weighted negatives}: For each discharge summary, we sampled candidate medical concepts according to their prevalence in MIMIC-IV. To characterize their difficulty, we additionally recorded the number of hops in the ICD hierarchy between each sampled negative and the closest related concept.

\textbf{Semantically similar negatives}: To increase semantic confusion, we embedded medical concept name and sampled the most similar but unrelated concept. For example, Obesity and BMI 30–39 both capture obesity-related conditions but belong to different categories in the ICD hierarchy (disease vs. BMI classification). These hard negatives are designed to test fine-grained reasoning about whether a concept is truly documented in the note.

Together, these strategies minimize the presence of trivial negatives and enrich the dataset with medically challenging contrasts.

\subsection{Annotation Candidate Classes}



We recruited 9 independent expert annotators with medical training to serve as the final authority. Two annotators were presented with a discharge summary and a candidate medical concept and asked to label the pair as Related or Unrelated. For Related cases, annotators highlighted evidence spans (character offsets) in the note and provided a short textual justification. Only cases where both annotators agreed were retained in the dataset.

To ensure label quality, all annotations underwent a two-stage review:

\begin{itemize}
  \item Initial Review – An automated program compared the two independent annotations for each note–concept pair and refined highlights as needed. The program flagged mismatches between classification (related / unrelated); these issues were resolved before proceeding.
  \item SME Review – A subject matter expert independently verified the selected label for correctness, completeness, and consistency, ensuring evidences were medically sound.
\end{itemize}

We recruited 9 independent expert annotators with medical training. For each note and concept pair, two annotators independently reviewed the discharge summary and labeled the concept as \textit{Related} or \textit{Unrelated}. For \textit{Related} pairs, annotators additionally highlighted sentence-level evidence spans (recorded as character offsets) and provided a short clinical justification.

Disagreements between the two annotators were not included without adjudication. Specifically, if the initial labels disagreed (\textit{Related} vs.\ \textit{Unrelated}), the example was flagged for resolution and reviewed to determine a single final label; pairs that could not be resolved to a clear final decision were excluded from the release. The final dataset therefore contains only examples with a verified gold label and accompanying evidence when applicable.

Through this pipeline, we curated a final dataset of 352 gold positives (explicit/implicit) and 471 gold negatives (unrelated). The resulting dataset emphasizes implicit reasoning, semantic ambiguity, and LLM disagreement, providing a challenging benchmark for medical concept extraction models.

\section{Evaluation metrics}

To evaluate performance on this dataset, we define two subtasks. 

(A) Medical Concept Extraction: Given a medical note and a candidate medical concept, the system predicts whether the note documents that concept. We compute micro-averaged Precision, Recall, and F1 across all cases, where a true positive is defined as predicting True when the gold label is True. 

(B) Sentence-Level Evidence Retrieval: For cases with expert-annotated evidence, we measure how well a system retrieves the correct sentences. We report macro-averaged sentence recall, defined as the proportion of gold evidence sentences correctly retrieved. If no sentences are returned for a case with gold evidence, then the recall for this case is 0. 

Together, these metrics assess both the correctness of concept extraction and the interpretability of the reasoning process.

\begin{figure}[h]
\begin{center}
\includegraphics[width=0.95\textwidth]{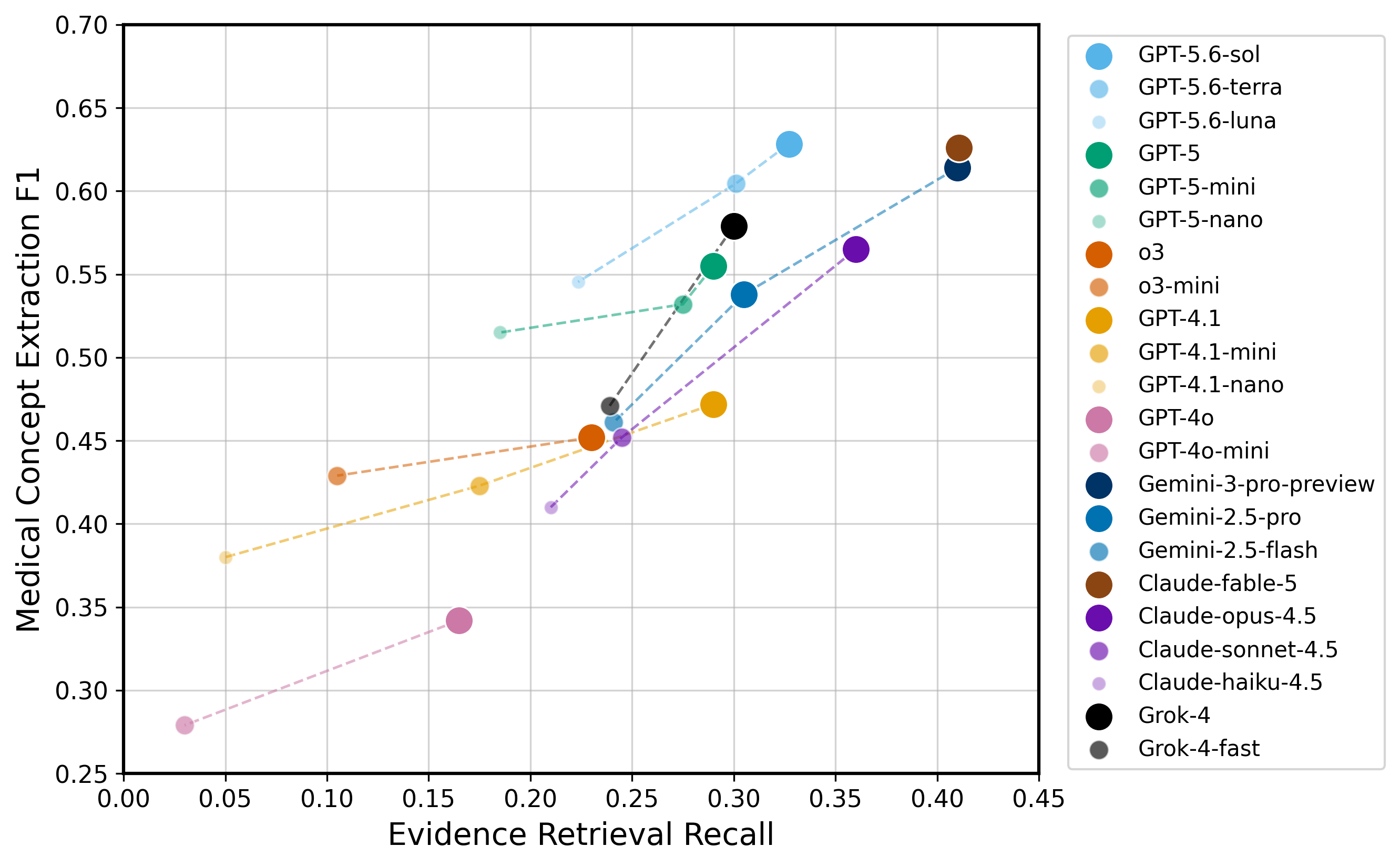}
\end{center}
\caption{Performance of LLMs on the medical concept extraction task (y-axis) and evidence retrieval task (x-axis).
}
\label{fig:relation}
\end{figure}

\section{Results}
We report MedicalBench results for different models and analyze performance by 
concept extraction (Section \ref{sec:result1}), 
evidence retrieval (Section \ref{sec:result2}), 
relationship between concepts and evidences (Section \ref{sec:result3}), 
the need for implicit prompt (Section \ref{sec:result4}),
relationship between extraction and note length (Section \ref{sec:lengthbias}),
and case studies (Section \ref{sec:casestudy}).

\subsection{Medical Concept Extraction}
\label{sec:result1}
\begin{figure}[h]
\begin{center}
\includegraphics[width=0.95\textwidth]{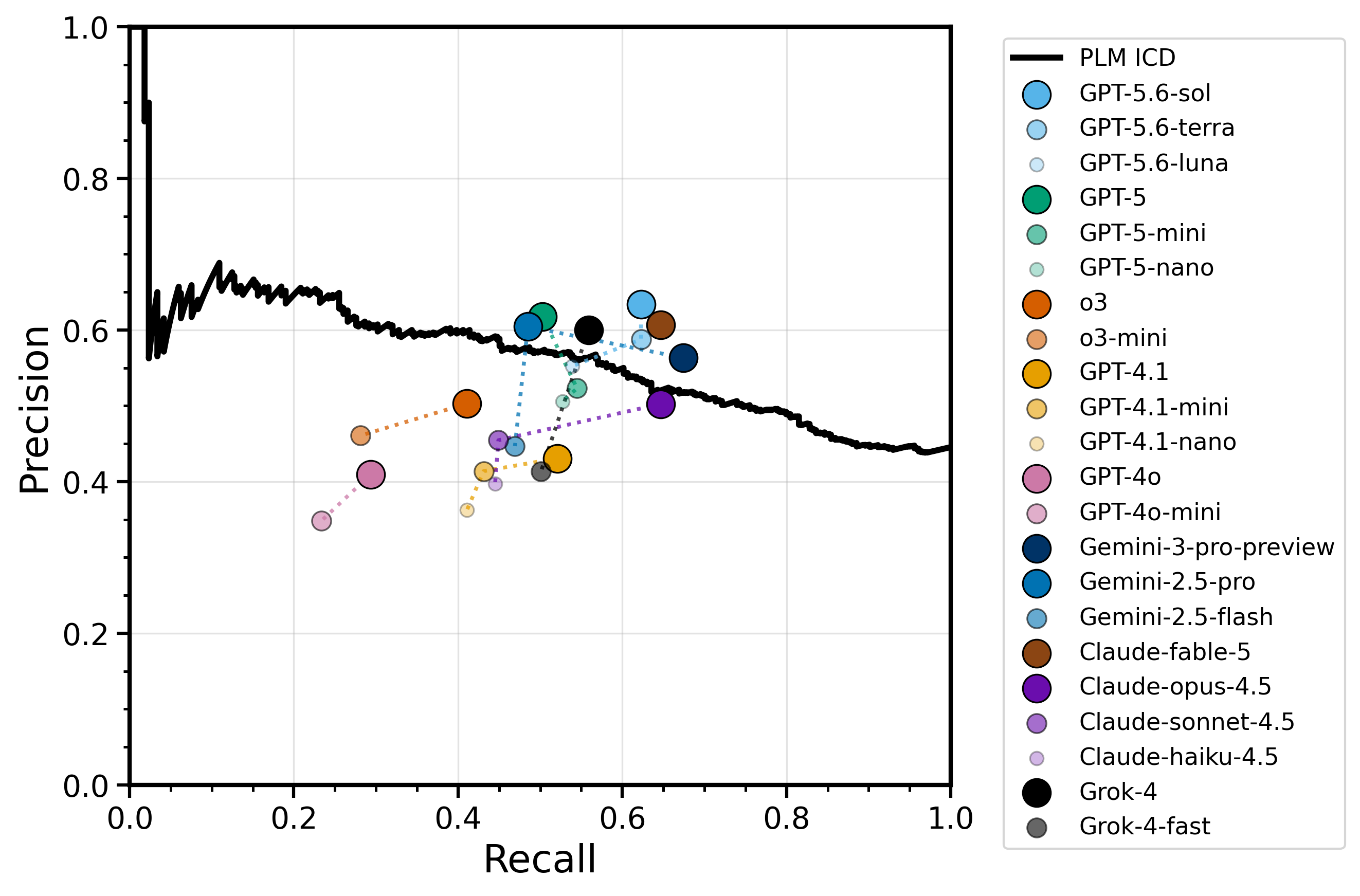}
\end{center}
\caption{Precision–recall comparison of LLMs and a pretrained language model (PLM-ICD), which is an encoder with $\approx 125M$ parameters, on the medical concept extraction task.
}
\label{fig:prcurve}
\end{figure}

To assess medical concept extractions, we evaluate each model’s ability to judge whether medical note documents the candidate medical concept. The y-axis of Figure \ref{fig:relation} shows medical concept extraction F1. GPT-5.6-sol and Claude-fable-5 achieve the best F1 score of 0.62, while Gemini-3-pro-preview, Grok-4, Claude-opus-4.5, GPT-5 achieve F1 scores of 0.61, 0.57, 0.56, and 0.55. The earlier reasoning model (O3) achieves F1 of 0.45 and non-reasoning model (GPT-4.1) achieves 0.47. Models with more parameters perform better. For example, Claude-opus-4.5 (0.56) outperforms its sonnet version (0.45) and haiku version (0.41). Across all systems, F1 scores remain below 0.70, reflecting the inherent difficulty of MedicalBench, which requires nuanced understanding of implicit disease mentions and fine-grained reasoning over long notes.

Figure \ref{fig:prcurve} presents a precision–recall comparison between zero-shot LLMs and a supervised pretrained language model baseline fine-tuned for medical concept extraction (PLM-ICD).
Although PLM-ICD provides a strong discriminative baseline, several recent LLMs operate above its precision–recall curve. GPT-5.6-sol and Claude-fable-5 achieve the strongest overall balance, combining recall above 0.60 with precision of approximately 0.61–0.63. Gemini-3-pro-preview attains the highest recall, approximately 0.67, while maintaining a precision of about 0.56. GPT-5, Grok-4, Gemini-2.5-pro, and GPT-5.6-terra also surpass the PLM-ICD frontier at their respective operating points. In contrast, o3 and the GPT-4.1, GPT-4o, and smaller Claude variants generally exhibit less favorable precision–recall trade-offs. Overall, the results demonstrate that the strongest contemporary LLMs can match or outperform a task-specific supervised baseline without fine-tuning. The PLM-ICD area under the precision–recall curve of 0.55 further indicates that MedicalBench remains challenging for conventional BERT-based methods.

\subsection{Evidence Retrieval}
\label{sec:result2}
To assess interpretability, we evaluate each model’s ability to retrieve the sentences that support its medical concept extraction (X-axis of Figure \ref{fig:relation}). Claude-fable-5 again leads, achieving the highest evidence recall, followed by Gemini-3-pro-preview, Claude-opus-4.5, GPT-5.6-sol, and Gemini-2.5-pro, which also retrieve relevant evidence with strong precision. Models with smaller parameter counts (e.g., GPT-5-nano, GPT-4.1-nano) display diminished evidence recall, consistent with their weaker text comprehension and contextual grounding.

\subsection{Improved Medical Concept Extraction via Evidence Retrieval cues}
\label{sec:result3}
From Figure \ref{fig:relation}, models that better retrieve supporting evidence also tend to make more correct medical concept extractions, suggesting that evidence benefits concept extraction.

To further examine this link, we paraphrased the retrieved evidence into reasoning cues (e.g., ``The note describes a low hemoglobin level, which indicates anemia.'') and incorporated these statements directly into the prompt and then re-evaluated the medical concept extraction task. Table \ref{tab:resultwithreason} compares GPT-5’s performance with and without the inclusion of such reasononing cues.

\begin{table}[h]
\centering
\begin{tabular}{|l|c|c|c|c|c|c|c|}
\hline
\textbf{Condition} & \textbf{Precision} & \textbf{Recall} & \textbf{F1} & \textbf{TP} & \textbf{TN} & \textbf{FP} & \textbf{FN} \\
\hline
GPT-5 Without Reasoning Cues & 0.6317 & 0.4971 & 0.5564 & 175 & 369 & 102 & 177 \\
GPT-5 With Reasoning Cues   & 0.7578 & 0.6845 & 0.7194 & 241 & 394 & 77  & 111 \\
\hline
\end{tabular}
\caption{GPT-5 Performance metrics comparison with and without reasoning cues}
\label{tab:resultwithreason}
\end{table}

Providing reasoning cues leads to a substantial performance gain: the F1 score increases from 0.55 to 0.71, driven by improvements in both precision (+12 points) and recall (+18 points). This improvement suggests that explicitly structured reasoning cues act as a bridge between retrieval and decision-making, helping the model not only identify concept but also interpret meaningful cues. Moreover, the reduction in false positives and false negatives indicates enhanced discrimination between true and confusable concept candidates.
These results demonstrate that integrating human-interpretable reasoning cues into prompts can meaningfully strengthen extraction quality and highlight a promising direction for explainable medical LLMs.

\subsection{Improved medical concept extraction via extraction of implicit evidence}
\label{sec:result4}
\label{sec:naive}
We next investigated the role of explicit mentions of extracting implicit evidence in the prompt,
by ablating the prompt for the model to consider not only explicit disease mentions but also implicit expressions of conditions (e.g., BMI 37 may indicate obesity). To quantify its contribution, we remove the extraction of implicit evidence from the prompt (Section \ref{sec:prompt}). The results are shown in Table \ref{tab:resultwithimplicit}.

\begin{table}[h]
\centering
\begin{tabular}{|l|c|c|c|c|c|c|c|}
\hline
\textbf{Condition} & \textbf{Precision} & \textbf{Recall} & \textbf{F1} & \textbf{TP} & \textbf{TN} & \textbf{FP} & \textbf{FN} \\
\hline
GPT-5 With Implicit Evidence & 0.6317 & 0.4971 & 0.5564 & 175 & 369 & 102 & 177 \\
GPT-5 Without Implicit Evidence & 0.4970 & 0.2414 & 0.3250 & 85 & 385 & 86  & 267 \\
\hline
\end{tabular}
\caption{GPT-5 Performance metrics comparison with and without extraction of implicit evidence}
\label{tab:resultwithimplicit}
\end{table}

When the implicit evidence extraction is removed from the prompt, GPT-5’s performance declines substantially across all metrics: F1 drops from 0.55 to 0.32, precision decreases from 0.63 to 0.49, and recall nearly halves from 0.49 to 0.24. The model thus becomes both less accurate and less sensitive, failing to recognize many implicit positives while also producing more false alarms. This indicates that prompting the model to explicitly account for implicit evidence not only helps it capture hidden relationships between medical concepts but also sharpens its decision boundary, improving discrimination between true and false relations.

These results highlight that instructions encouraging inferential reasoning are essential for accurate medical concept extraction, as many diagnoses in medical notes are only implied rather than directly stated. Without guidance to infer such relationships, models lose both precision and recall—highlighting that implicit awareness contributes simultaneously to sensitivity and selectivity in medical reasoning.

\subsection{Medical concept extraction and note length}
\label{sec:lengthbias}
Length bias, where longer inputs reduce extraction accuracy, is well documented in text extraction.
Studies such as LongBench \citep{bai-etal-2024-longbench}, Lost in the Middle \citep{liu-etal-2024-lost}, and Long-Context LLMs Meet RAG \citep{Jin2024LongContextLM} have shown that as the context length grows, models become distracted by irrelevant evidence, leading to a lower extraction accuracy.  
In the medical domain prior work \citep{Ji2021DoesTM,Liu2022AutomatedIC} also reported that medical concept extraction is less precise for longer discharge summaries.

\begin{figure}[h]
\begin{center}
\includegraphics[width=0.95\textwidth]{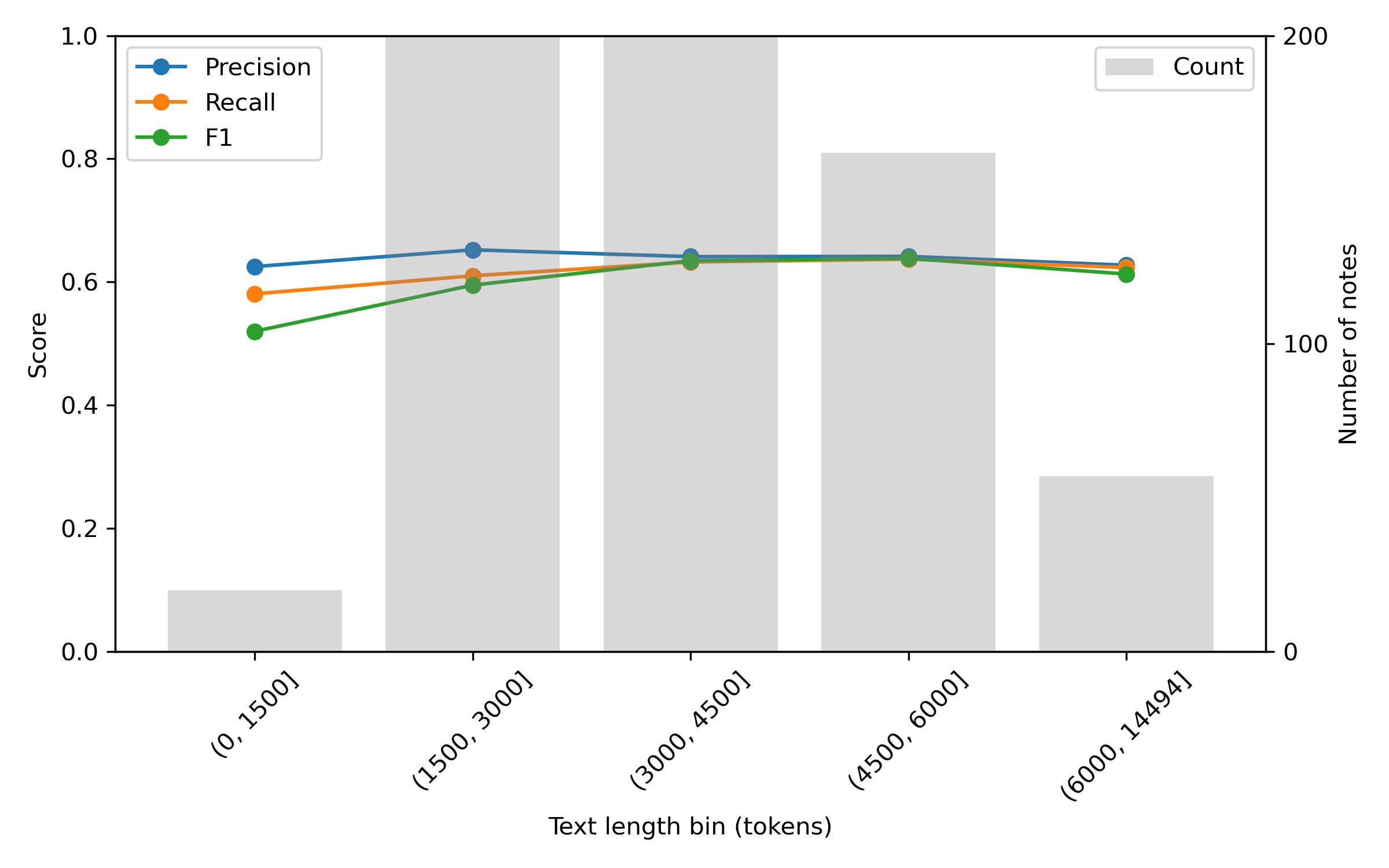}
\end{center}
\caption{GPT-5 performance across different note lengths}
\label{fig:lengthperf}
\end{figure}

However, MedicalBench does not exhibit such a note length bias.
As shown in Figure \ref{fig:lengthperf}, precision, recall, and F1 remain stable across note length bins, with no monotonic decline.
We also conducted an equivalence test for correlation using the Two One-Sided Tests (TOST) procedure with bound of 0.1. The observed correlation was $r=0.02$, and the 90\% confidence interval was $[-0.03, 0.07]$. Both one-sided tests were significant (TOST p-value = 0.0109), supporting equivalence. Thus, the correlation between note length and accuracy is statistically negligible within the predefined bounds.
This suggests that note length explains little of the variance in performance compared with other factors such as implicit reasoning.
This finding indicates that the MedicalBench design effectively mitigates confounding effects of note length and isolates reasoning difficulty as the main challenge.

\subsection{Case studies}
\label{sec:casestudy}
To further illustrate the reasoning challenges posed by MedicalBench, we present three representative examples highlighting \textit{implicit medical reasoning}, \textit{temporal context}, and \textit{contextual misinterpretation}. These cases demonstrate how even high-performing models such as GPT-5 can make plausible but incorrect decisions, underscoring the dataset’s difficulty and the need for medically grounded interpretability.

\subsubsection{Case Study 1: Implicit Inference from Medication Patterns}

\noindent\textbf{Medical Concept:} \textit{Unspecified kidney failure} \\
\textbf{Medical Note (excerpt):}
\begin{quote}
\small
\textit{calcium acetate 667 mg tablet — 2 tabs orally 3 times a day (with meals), Status: Active} \\
\textit{valganciclovir 450 mg tablet — 1 tab orally 3 times weekly (Mon/Wed/Fri), Status: Active} \\
\textit{furosemide 80 mg tablet — 1 tab orally twice daily, Status: Active}
\end{quote}

\noindent\textbf{Model Output:} GPT-5 predicted that the medical concept is \textbf{not related} to the note, reasoning that kidney failure was not explicitly mentioned.  

\noindent\textbf{Expert Judgment:} The concept is \textbf{related}. Several medications in the note strongly imply chronic kidney disease or renal failure:
First, Valganciclovir is an antiviral  prescribed to renal transplant or advanced kidney disease patients.
Second, Furosemide is a loop diuretic used to treat fluid overload in chronic kidney failure.
Third, Calcium acetate is a phosphate binder routinely used in dialysis patients \citep{Limaye2023LetermovirCMV}.

These pharmacological cues collectively suggest significant renal dysfunction, even in the absence of an explicit diagnosis. This case underscores the need for implicit reasoning and domain knowledge integration, requiring models to infer conditions from treatment patterns rather than literal text.

\subsubsection{Case Study 2: Misinterpretation of Temporal Context}

\noindent\textbf{Medical Concept:} \textit{Old myocardial infarction} \\
\textbf{Medical Note (excerpt):}
\begin{quote}
\small
\textit{IMPRESSION: 1) Severe regional in setting of moderate global LV systolic dysfunction consistent with Takotsubo's cardiomyopathy; however, myocardial infarction in proximal LAD cannot be excluded, though less likely given distribution of regionality.}
\end{quote}

\noindent\textbf{Model Output:} GPT-5 predicted that the medical concept is \textbf{related} to the note, citing the explicit mention of ``myocardial infarction''.  

\noindent\textbf{Expert Judgment:} The concept is \textbf{not related}. The phrase ``old myocardial infarction'' refers specifically to a \textit{previous (remote, healed)} infarction rather than a current or suspected acute event. The medical note describes an ongoing evaluation for \textit{acute chest pain} with differential diagnoses of NSTEMI versus Takotsubo’s cardiomyopathy. Importantly, the note states that myocardial infarction was \textit{not confirmed} and was considered \textit{less likely}. No evidence supports a prior infarction, which is required for the concept.  

This case highlights the challenge of temporal reasoning—distinguishing \textit{past} versus \textit{current} disease states—and shows that even advanced models may conflate lexical overlap (``myocardial infarction'') with valid evidence for medical concept extractions.

\subsubsection{Case Study 3: Scattered Evidences}

\noindent\textbf{Medical Concept:} \textit{Excision of Thoracic Vertebra, Percutaneous Approach, Diagnostic} \\
\textbf{Medical Note (excerpt):}
\begin{quote}
\small
\textit{Major Surgical or Invasive Procedure:
T11/L1 kyphoplasty
… (1446 tokens later) …
Assessment/Plan: 
[] F/u pathology results: bone biopsy of the L1 and T11 vertebral bodies.
[] Home metoprolol increased to 12.5mg BID given afib with RVR on admission. Please ensure cardiology follow up as an outpatient.}
\end{quote}

\noindent\textbf{Model Output:} GPT-5 predicted that the concept is \textbf{not related} to the note.  

\noindent\textbf{Expert Judgment:}
The concept is \textbf{related}. There are two evidences in this note. First, ``bone biopsy of the L1 and T11 vertebral bodies'' documents that tissue from T11 (a thoracic vertebra) was obtained for diagnostic purposes, establishing an excisional biopsy at the correct anatomical site and with diagnostic intent. Without this, there is no support for an \textit{excision/biopsy} of a thoracic vertebra. Second, the note also records ``T11/L1 kyphoplasty'' in another section of the note. This indicates a \textit{percutaneous} procedure typically performed via percutaneous transpedicular access in the same operative session. Without this, there is no support for percutaneous approach.

Both evidences are essential and complementary: Evidence 1 supplies diagnostic excision at the thoracic site (T11), and evidence 2 supplies the percutaneous approach. If either is missing, the concept extraction fails—either the action at T11 is not proven (no diagnostic excision) or the approach is not supported (no percutaneous access). The long-range separation of these evidences (1,446 tokens apart) illustrates a scattered evidence failure case, where models must retrieve and integrate distant, cross-referential details to correctly extract medical concpets.

All examples above reveal fundamental aspects that make MedicalBench difficult:
\begin{itemize}[leftmargin=2em]
    \item \textbf{Ambiguous evidences}, where correctness depends on nuanced interpretation rather than explicit mention.
    \item \textbf{Subtle temporal distinctions}, such as differentiating old versus acute disease states.
    \item \textbf{Scattered evidences}, where essential evidences are scattered in the medical notes.
\end{itemize}

Such cases explain why model F1 scores remain below 0.7: the dataset intentionally tests beyond surface-level keyword recognition. Future models that integrate temporal understanding, pharmacological reasoning, and evidence-aware explanations may better capture these nuanced relationships essential for trustworthy medical concept extractions.

\section{Conclusion}
This work introduces MedicalBench, a new benchmark designed to evaluate not only the correctness but also the reasoning transparency of automated medical concept extractions systems. Unlike existing datasets that emphasize binary concept extraction, MedicalBench incorporates implicit and confusable diagnostic cases, along with expert-verified evidence spans, providing the first large-scale resource for assessing why a medical concept should be extracted.

Our experiments across a range of LLMs demonstrate that MedicalBench is a challenging benchmark (best F1 score $<$ 0.7), requiring nuanced comprehension and reasoning beyond surface-level pattern matching. Models that effectively retrieve evidence and articulate reasoning achieve markedly higher concept extractions accuracy, revealing a clear link between interpretability and performance. Incorporating explicit reasoning cues or implicit evidence extraction into prompts further enhances outcomes, underscoring the need for models that can reason medically rather than merely classify text. Finally, MedicalBench seeks to transition the field from extraction-as-classification toward extraction-as-reasoning, a shift essential for deploying large language models responsibly in real-world medical settings.

\section{Acknowledgment}
This project is built on de-identified clinical data from the MIMIC-IV database, which has been publicly released for research under a data use agreement. All patient health information in the discharge summaries was de-identified in accordance with HIPAA safe harbor standards by the original MIMIC-IV data curators. The annotations we added (evidence and explanations) contain no patient-identifying information, only medical insights. Institutional Review Board (IRB) approval was not required for this work, as it does not involve identifiable human subjects. We thank Dr. Deqi Kong for clinical input, Jiafeng Song and Sepehr Janghorbani for feedback on evaluation and Elizabeth Erickson and Kalyan Guntupalli for their administrative support. We are grateful to the MIT Laboratory for Computational Physiology and PhysioNet for providing access to the MIMIC-IV databases used in this study.

\bibliography{iclr2025_conference}
\bibliographystyle{iclr2025_conference}

\appendix
\section{Appendix}

\subsection{Prompt template}
\label{sec:prompt}

Below is the template to extract medical concept and locate evidence. Unless otherwise specified, it is the default prompt for all LLMs in our experiments. 

\begin{tcolorbox}[colback=gray!5!white, colframe=blue!75!black, title=Prompt to extract concept and locate evidence, sharp corners, boxrule=0.8pt]
You are an expert in annotating clinical notes. Your task is to identify the type of mention for a given medical concept within a provided clinical note. There are three possible mention types:

- Explicit: entity that is found directly in the document. For example, the discharge diagnosis includes 'Acute Kidney Injury', which corresponds to the acute kidney failure with tubular necrosis entity.

- Implicit: entity which is not directly mentioned in the document but can be inferred using context. For example, The mention of low hemoglobin level suggests anemia.

- Unrelated: entity which is not related to the clinical note.

Output format: A json dictionary, where the json is of the following keys

`extracted mention`: List[str], extracted mentions for medical concept from clinical note.

`mention type`: str, either "Explicit", "Implicit", or "Unrelated".

Medical concept:
\{label\}

Clinical note: 
\{text\}
\end{tcolorbox}

Below is the template for the ablation study where we remove the categories from the prompt (Section \ref{sec:naive}).

\begin{tcolorbox}[colback=gray!5!white, colframe=blue!75!black, title=Prompt to extract concept and locate evidence without categories, sharp corners, boxrule=0.8pt]
You are an expert in annotating clinical notes. Your task is to identify the medical concept within a provided clinical note. 

Output format: A json dictionary, where the json is of the following keys.

`extracted mention`: List[str], extracted mentions for medical concept from clinical note. If no evidence, then return empty.

`is related`: bool, If medical concept is related to the clinical note, return true, otherwise return false.

Medical concept:
\{label\}

Clinical note: 
\{text\}
\end{tcolorbox}

\end{document}